\documentclass{svproc}
\usepackage{graphicx}%
\usepackage{multirow}%
\usepackage{amsmath,amssymb,amsfonts}%
\usepackage{mathrsfs}%
\usepackage[title]{appendix}%
\usepackage{xcolor}%
\usepackage{textcomp}%
\usepackage{manyfoot}%
\usepackage{booktabs}%
\usepackage{algorithm}%
\usepackage{algorithmicx}%
\usepackage{algpseudocode}%
\usepackage{listings}%
\usepackage{graphicx}%
\usepackage{multirow}%
\usepackage{mathrsfs}%
\usepackage[title]{appendix}%
\usepackage{xcolor}%
\usepackage{textcomp}%
\usepackage{subfig}
\usepackage{comment}
\usepackage{bbm}
\usepackage{paralist}
\usepackage{wrapfig}
\usepackage{url}

\begin{document}
\mainmatter              
\title{Heterophily-Based Graph Neural Network for Imbalanced Classification}
\titlerunning{Heterophily-Based Graph Neural Network for Imbalanced Classification}  
%
\author{Zirui Liang\inst{1} \and Yuntao Li\inst{1} \and
Tianjin Huang\inst{1} \and Akrati Saxena\inst{2} \and Yulong Pei\inst{1} \and Mykola Pechenizkiy\inst{1}}
\authorrunning{Zirui Liang et al.} 

\institute{Eindhoven University of Technology, Eindhoven, The Netherlands\\
\email{\{y.pei.1, m.pechenizkiy\}@tue.nl}
\and
Leiden Institute of Advanced Computer Science, \\Leiden University, Leiden, The Netherlands\\
\email{a.saxena@liacs.leidenuniv.nl}
}

\maketitle              

\begin{abstract}
Graph neural networks (GNNs) have shown promise in addressing graph-related problems, including node classification. However, in real-world scenarios, data often exhibits an imbalanced, sometimes highly-skewed, distribution with dominant classes representing the majority,
where certain classes are severely underrepresented. This leads to a suboptimal performance of standard GNNs on imbalanced graphs. In this paper, we introduce a unique approach that tackles imbalanced classification on graphs by considering graph heterophily. We investigate the intricate relationship between class imbalance and graph heterophily, revealing that minority classes not only exhibit a scarcity of samples but also manifest lower levels of homophily, facilitating the propagation of erroneous information among neighboring nodes. Drawing upon this insight, we propose an efficient method, called \texttt{Fast Im-GBK}, which integrates an imbalance classification strategy with heterophily-aware GNNs to effectively address the class imbalance problem while significantly reducing training time. Our experiments on real-world graphs demonstrate our model's superiority in classification performance and efficiency for node classification tasks compared to existing baselines.
\end{abstract}
\section{Introduction}\label{sec:intro}
GNNs have gained popularity for their accuracy in handling graph data. However, their accuracy, like other deep learning models,  is highly dependent on data quality. One major challenge is class imbalance, where some classes have far fewer examples than others. This can lead to biased classification results, favoring the majority class while neglecting the minority classes \cite{japkowicz2002class}. The issue of imbalanced datasets commonly arises in classification and recognition tasks, where accurate classification of minority classes is critical. Graph imbalance classification has real-world applications, like identifying spammers in social networks \cite{wu2020graph} and detecting fraud in financial networks \cite{liu2021pick}. In these cases, abnormal nodes are rare, making graph imbalance classification very challenging. Finding effective solutions to this problem is valuable for both research and practical applications. 

The class-imbalanced problem has been extensively studied in machine learning and deep learning, as evident by prior research \cite{johnson2019survey}. However, these methods may not effectively handle imbalanced graph data due to the interconnected nature of nodes within graphs. Graph nodes are characterized not only by their own properties but also by the properties of their neighboring nodes, introducing non-i.i.d. (independent and identically distributed) characteristics. Recent studies on graph imbalance classification have focused on data augmentation techniques, such as GraphSMOTE \cite{zhao2021graphsmote} and GraphENS \cite{park2021graphens}. However, our observations indicate that class imbalance in graphs is often accompanied by heterophilic connections of minority nodes, where minority nodes have more connections with nodes of diverse labels than the majority class nodes. This finding suggests that traditional techniques may be insufficient in the presence of heterophily.

To address this challenge, we propose incorporating a graph heterophily handling strategy into graph imbalanced classification. Our approach builds upon the bi-kernel design of GBK-GNN \cite{2021GBK} to capture both homophily and heterophily within the graph. Additionally, we introduce a class imbalance-aware loss function, such as logit adjusted loss, to appropriately reweight minority and majority nodes. The complexity of GBK-GNN makes training computationally challenging. To overcome this, we propose an efficient version of the GBK-GNN that achieves both efficacy and efficiency in training.

Our main contributions are as follows:
(1) We provide comprehensive insights into the imbalance classification problem in graphs from the perspective of graph heterophily and investigate the relationship between class imbalance and heterophily. (2) We present a novel framework that integrates graph heterophily and class-imbalance handling based on the insights and its fast implementation that significantly reduces training time. 
(3) We conduct extensive experiments on various real-world graphs to validate the effectiveness and efficiency of our proposed framework in addressing imbalanced classification on graphs.

\section{Related Work}\label{rw}

\par \textbf{Imbalanced Classification}\label{ic}
Efforts to counter class imbalance in classification entail developing unbiased classifiers that account for label distribution in training data. Existing strategies fall into three categories: loss modification, post-hoc correction, and re-sampling techniques. Loss modification adjusts the objective function by assigning greater weights \cite{japkowicz2002class} to minority classes. Post-hoc correction methods \cite{longtail} adapt logits during inference to rectify underrepresented minority class predictions. Re-sampling employs techniques, such as sampling strategies \cite{ren2020balanced} or data generation \cite{chawla2002smote}, to augment minority class data. The widely utilized Synthetic Minority Over-sampling Technique (SMOTE) \cite{chawla2002smote} generates new instances by merging minority class data with nearest neighbors. 

To tackle class imbalance in graph-based classification, diverse approaches harness graph structural information to mitigate the challenge. GraphSMOTE \cite{zhao2021graphsmote} synthesizes minor nodes by interpolating existing minority nodes, with connectivity guided by a pretrained edge predictor. The Topology-Aware Margin (TAM) loss \cite{song2022tam} considers each node's local topology by comparing its connectivity pattern to the class-averaged counterpart. When nearby nodes in the target class are denser, the margin for that class decreases. This change enhances learning adaptability and effectiveness through comparison.
GraphENS \cite{park2021graphens} is another technique that generates an entire ego network for the minor class by amalgamating distinct ego networks based on similarity. These methods effectively combat class imbalance in graph-based classification, leveraging graph structures and introducing inventive augmentation techniques.

\par \textbf{Heterophily Problem} 
In graphs, homophily \cite{mcpherson2001birds} suggests that nodes with similar features tend to be connected, and heterophily suggests that nodes with diverse features and class labels tend to be connected. Recent investigations have analyzed the impact of heterophily on various tasks, emphasizing the significance of accounting for attribute information and devising methods attuned to heterophily \cite{zheng2022graph}. Newly proposed models addressing heterophily can be classified into two categories: non-local neighbor extension and GNN architecture refinement \cite{zheng2022graph}. Methods grounded in non-local neighbor extension seek to alleviate this challenge through neighborhood exploration. The H2GCN method \cite{zhu2020beyond}, for instance, integrates insights from higher-order neighbors, revealing that two-hop neighbors frequently encompass more nodes of the same class as the central ego node. NLGNN \cite{liu2022towards} follows a pathway of employing attention mechanisms or pointer networks to prioritize prospective neighbor nodes based on attention scores or their relevance to the ego node. Approaches enhancing GNN architectures aspire to harness both local and non-local neighbor insights, boosting model capacity by fostering distinct and discerning node representations. A representative work is GBK-GNN (Gated Bi-Kernel Graph Neural Network) \cite{2021GBK}, which employs dual kernels to encapsulate homophily and heterophily details and introduces a gate to ascertain the kernel suitable for a given node pair.

\section{Motivation}\label{mt}

Node classification on graphs, such as one performed by the Graph Convolutional Network (GCN), differs fundamentally from non-graph tasks due to the interconnectivity of nodes.
In imbalanced class distributions, minority nodes may have a higher proportion of heterophilic edges in their local neighborhoods, which can negatively impact classification performance.

To investigate the relationship between homophily and different classes, especially minorities, we conducted a small analysis on four datasets: Cora, CiteSeer, Wiki, and Coauthor CS (details about datasets can be found in Section \ref{expdata}). Our analysis involves computing the average homophily ratios and calculating node numbers across different categories. In particular, the average homophily ratio for nodes with label $y$ is defined as:
\begin{equation}
\small
h\left(y,\mathcal{G}_{\mathcal{V}}\right)=\frac{1}{|\mathcal{V}_{y}|} \sum_{i \in \mathcal{V}_{y}} \frac{|\mathcal{N}_{i^s}|}{|\mathcal{N}_{i}|}
\end{equation}
where $\mathcal{V}{y}$ represents a set of nodes with label $y$, $\mathcal{N}{i}$ is the set of neighbors of node ${v_i}$ (excluding ${v_i}$) in graph $\mathcal{G}$, and $\mathcal{N}_{i^s}$ is the set of neighbors (excluding ${v_i}$) whose class is the same as ${v}_{i}$.

\begin{figure}[t]
\centering
\subfloat[Distribution of nodes in Cora.]{\includegraphics[width=.4\textwidth]{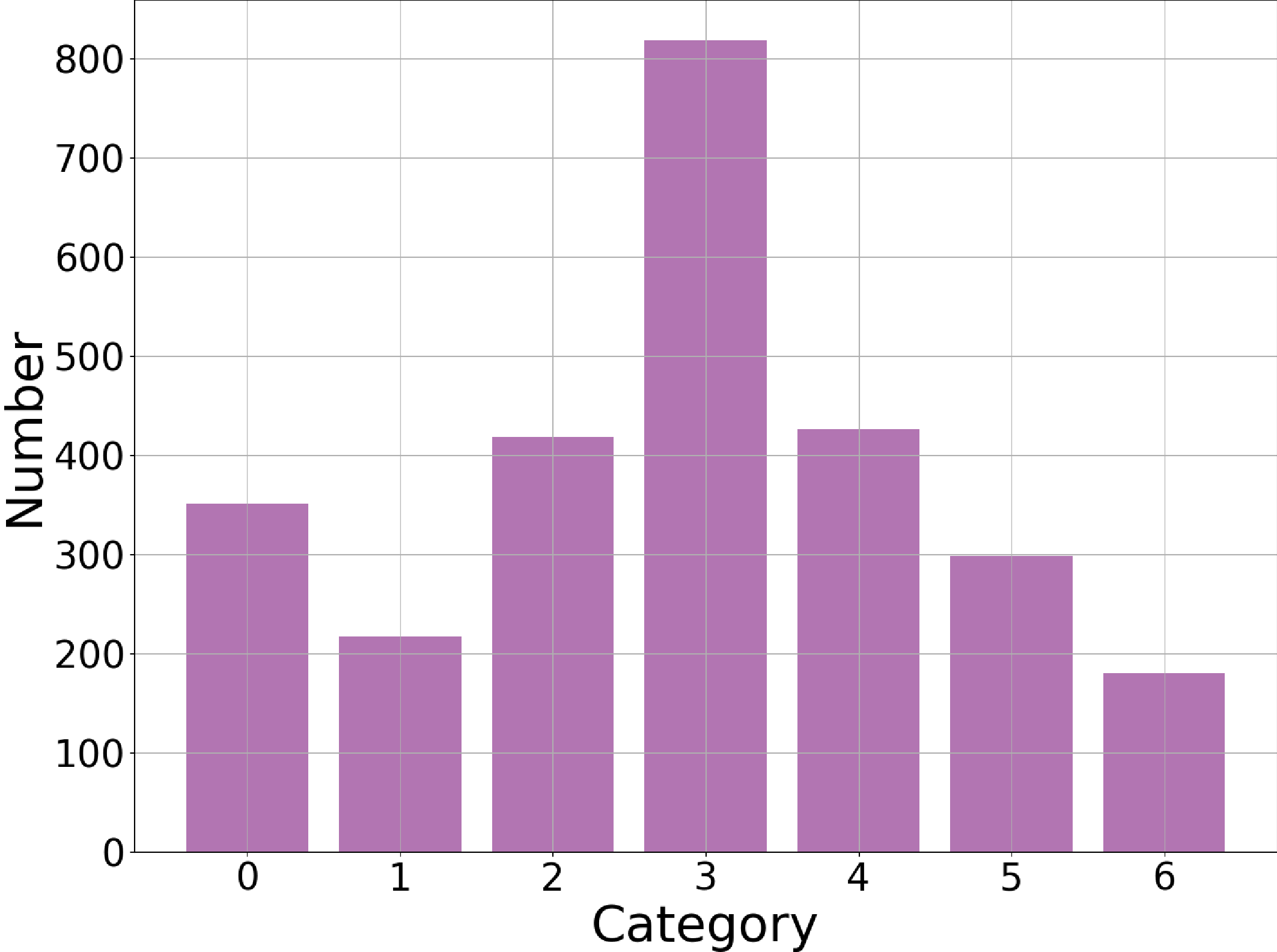}} \qquad
\subfloat[Homophily Ratio in Cora.]{\includegraphics[width=.4\textwidth]{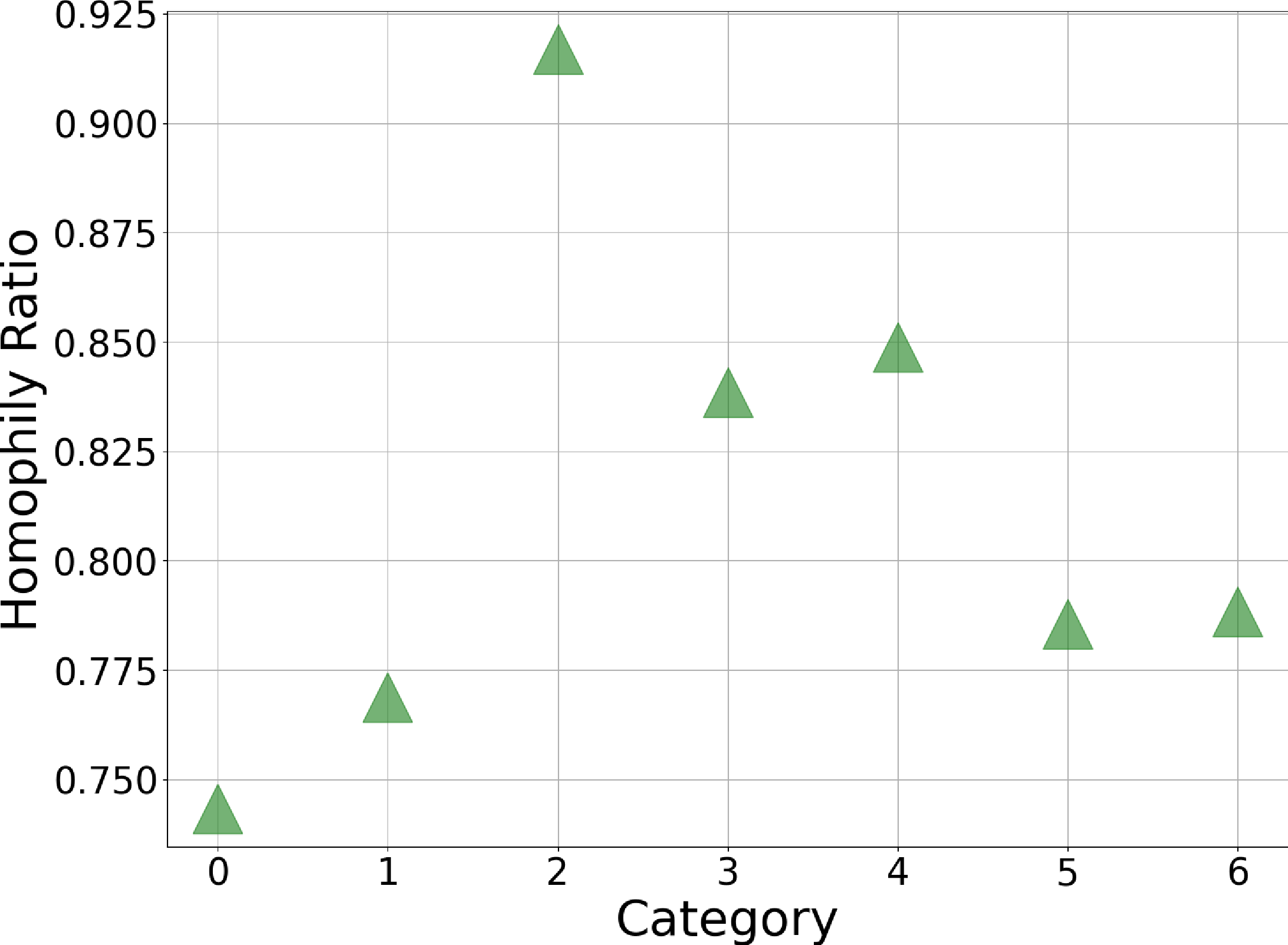}} \\
\caption{Category distributions (left) and average homophily ratios (right) of Cora.}
    \label{analysisP}
\end{figure}

\paragraph{\textit{Observations.}} The results for Cora datasets are shown in Fig. \ref{analysisP}. It is evident that the average homophily ratios of minority classes are relatively smallersuggesting higher proportions of heterophilic edges in their local neighborhoods. This could negatively affect classification performance when using existing imbalance strategies like data augmentation and loss reweight. This finding highlights the importance of considering the homophily and heterophily properties of nodes when designing graph classification models. We propose a novel approach that addresses imbalance issue effectively, considering node heterophily. 

\paragraph{\textit{Problem Formulation.}} An attributed graph is denoted by $\mathcal{G}=(\mathcal{V}, \mathcal{E}, \mathrm{X})$, where $\mathcal{V} = \left\{v_1, \ldots, v_n\right\}$ represents the set of $n$ nodes in $\mathcal{G}$, and $\mathcal{E}$ is the set of edges with edge $e_{ij}$ connecting nodes $v_i$ and $v_j$. For simplicity, we consider undirected graphs, but the argument could be generalized for directed graphs. The node attribute matrix is represented by $\mathrm{X}=\left[\mathbf{x}_1^{\top}, \ldots, \mathbf{x}_n^{\top}\right] \in \mathcal{R}^{n \times d}$, where $\mathbf{x}_i$ indicates the feature of node $v_i$. $\mathcal{Y}$ denotes the label information for nodes in $\mathcal{G}$ and each node $v_i$ is associated with a label $y_i \in \mathcal{R}^{n}$. During the training, only a subset of the classes, denoted by $\mathcal{Y_L}$, is available containing the labels for node subset $\mathcal{V_L}$. Moreover, $\mathcal{C} = \{c_1, c_2,\ldots,c_m\}$ represents the set of $m$ classes assigned to each node, with $|C_i|$ denoting the size of the $i$-th class, which indicates the number of nodes with class $c_i$. To quantify the degree of imbalance in the graph, we use the imbalance ratio, defined as $\mathbf{r} = \frac{\max _i\left(\left|C_i\right|\right)}{\min _i\left(\left|C_i\right|\right)}$. For imbalanced graphs, the imbalance ratio of $\mathcal{Y_L}$ is high (Table \ref{Statistics} of Section \ref{secexp} shows that the first two datasets have relatively balanced classes and the last two have imbalanced classes). \\ 
\textbf{Imbalance Node Classification on Graphs.} In the context of an imbalanced graph $\mathcal{G}$ with a high imbalance ratio $\mathbf{r}$, the aim is to develop a node classifier $f: f(\mathcal{V}, \mathcal{E}, \mathrm{X}) \rightarrow \mathcal{Y}$ that can work well for classifying nodes belonging to both the majority and minority classes.

\section{Methodology}\label{meth}

In this section, we present our solution to address class imbalance that incorporates heterophily handling and imbalance handling components (Section \ref{imGBK}). We also propose a fast version that effectively reduces training time (Section \ref{Fast Im-GBK}). The main objective of our model is to minimize the loss of minority classes while ensuring accurate information exchange during the message-passing process.

\subsection{Im-GBK}
\label{imGBK}

\subsubsection{Heterophily Handling}
\label{m:HH}
We build our model based on the GBK-GNN \cite{2021GBK} model, which is a good model for graph classification, though not able to handle class imbalance. 
GBK-GNN is designed to address the lack of distinguishability in GNN, which stems primarily from the incapability to adjust weights adaptively for various node types based on their distinct homophily properties. As a consequence, a bi-kernel feature transformation method is employed to capture either homophily or heterophily information. In this work, we, therefore, introduce a learnable kernel-based selection gate that aims to distinguish if a pair of nodes are similar or not and then selectively choose appropriate kernels, i.e., homophily or heterophily kernel. The formal expression for the input transformation is presented below.
\begin{equation}
\small
\mathbf{h}_{i}^{(l)}=\sigma\left(\mathbf{W}_{\mathbf{f}} \mathbf{h}_{i}^{(l-1)}+\frac{1}{|\mathcal{N}(v_{i})|} \sum_{v_{j} \in \mathcal{N}\left(v_{i}\right)} \alpha_{i j} \mathbf{W}_{\mathbf{s}} \mathbf{h}_{j}^{(l-1)}+\left(1-\alpha_{i j}\right) \mathbf{W}_{\mathbf{d}} \mathbf{h}_{j}^{(l-1)}\right) \label{GBKProcess} 
\end{equation}
\begin{equation}
\small
    \alpha_{ij}=\textit{}{Sigmoid}\left(\mathbf{W}_{\mathbf{g}}\left[\mathbf{h}_{i}^{(l-1)}, \mathbf{h}_{j}^{(l-1)}\right]\right) 
    \label{GBK-Gate}
\end{equation}
\begin{equation} 
\small
\mathcal{L}=\mathcal{L}_{0} + \lambda \sum_{l}^{L} \mathcal{L}_{g}^{(l)} 
\label{logGBKloss}
\end{equation}
where $\mathbf{W}_{\mathbf{s}}$ and $\mathbf{W}_{\mathbf{d}}$ are the kernels for homophilic and heterophilic edges, respectively. The value of $\alpha_{ij}$ is determined by $\mathbf{W}_{\mathbf{g}}$ and the embedding layer of nodes $i$ and $j$. The loss function consists of two parts: $\mathcal{L}_{0}$, a cross-entropy loss for node classification, and $\mathcal{L}_{g}^{(l)}$, a label consistency-based cross-entropy, i.e., to discriminate if labels of a pair of nodes are consistent for each layer $l$ to guide the training of the selection gate or not. A hyper-parameter $\lambda$ is introduced to balance the two losses. The original GBK-GNN method does not explicitly address the class imbalance issue, which leads to the model being biased toward the majority class. Our method employs a class-imbalance awareness for the GBK-GNN design, and therefore it mitigates the bias.

\subsubsection{Class-Imbalance Handling with Logit Adjusted Loss}
\label{m:logit}
When traditional softmax is used on such imbalanced data, it can result in biases toward the majority class. This is because the loss function used to train the model typically treats all classes equally, regardless of their frequency. As a result, the model tends to optimize for overall accuracy by prioritizing the majority classes while performing poorly on the minority classes. This issue can be addressed by adjusting the logits (i.e., the inputs to the softmax function) for each class to be inversely proportional to their frequencies in the training data, which effectively reduces the weight of the majority classes and increases the weight of the minority classes \cite{longtail}. In this study, we calculate logit-adjusted loss as follows:
\begin{equation}
\small
\mathcal{L}_{logit-adjusted}=-\log \frac{e^{f_{y}(x)+ \tau \cdot \log \pi_{y}}}{\sum_{y^{\prime} \in[L]} e^{f_{y^{\prime}}(x)+ \tau \cdot \log \pi_{y^{\prime}}}}
\label{MylossPartA}
\end{equation}
where $\pi_{y}$ is the estimate of the class prior. In this approach, a label-dependent offset is added to each logit, which differs from the standard softmax cross-entropy approach. Additionally, the class prior offset is enforced during the learning of the logits rather than being applied post-hoc, as in other methods. 

\subsubsection{Class-Imbalance Handling with Balanced Softmax}
\label{m:BS}
Another approach called balanced softmax \cite{ren2020balanced} focuses on assigning larger weights to the minority classes and smaller weights to the majority class, which encourages the model to focus more on the underrepresented classes and improve their performance. The traditional softmax function treats all classes equally, which can result in a bias towards the majority class in imbalanced datasets. Balanced softmax, on the other hand, adjusts the temperature parameter of the softmax function to balance the importance of each class, effectively reducing the impact of the majority class and increasing the impact of the minority classes. Formally, given $N_k$, $l_{v}$, $y_{v}$, $l_{v, y_{v}}$ represent \textit{k-th} of class $N$, logit and the label of node $v$, logit associated with the true label $y_{v}$ for the input node $v$, respectively. In this study, the balanced softmax is calculated as:
\begin{equation}
\small
\mathcal{L}_{balanced-softmax}=-\log \frac{e^{l_{v, y_{v}} + \log N_{y_v}}}{\sum_{k\in \mathcal{Y}} e^{l_{v, k} + \log |N_k|}}.
\label{MylossPartA}
\end{equation}

\subsubsection{Loss Function}\label{losssubs}
We design our loss function to combine these two components that handle heterophily and class imbalance in the proposed Im-GBK model. The learning objective of our model consists of (i) reducing the weight of the majority classes and increasing the weight of the minority classes in the training data, and (ii) improving the model's ability to select the ideal gate. To achieve this objective, we incorporate two loss components into the loss function: 
\begin{equation}
\small
\mathcal{L}=\mathcal{L}_{im} + \lambda \sum_{l}^{L} \mathcal{L}_{g}^{(l)} \label{Myloss}.
\end{equation}
The first component, from the class-imbalance handler, denoted as $\mathcal{L}_{im}$, applies either the \textit{logit adjusted loss} $\mathcal{L}_{logit-adjusted}$ or the \textit{Balanced Softmax} $\mathcal{L}_{balanced-softmax}$ approach to reduce the impact of the majority class and focus on underrepresented classes. The second component, denoted as $\mathcal{L}{g}^{(l)}$, applies cross-entropy loss to each layer $l$ of the heterophily handler to improve the model's discriminative power and adaptively select gate to emphasize homophily or heterophily. The hyper-parameter $\lambda$ balances the two losses in the overall loss function.

\subsection{Fast Im-GBK}\label{Fast Im-GBK}

The major limitation of Im-GBK lies in efficiency, as the additional time is required to compute the gate result from the heterophily handling component. The challenge arises from the need to label edges connecting unknown nodes, which is typically addressed by learning an edge classifier. Specifically, in the message-passing process, the gate acts as an edge classifier that predicts the homophily level of a given edge. As a result, this process requires additional time proportional to the number of edges in the graph, expressed as $T_{\textbf{extra}} = |\mathbf{E}| \times time(e_{ij})$, where $time(e_{ij})$ represents the time to compute one edge. 
Therefore, we propose to use a graph-level homophily ratio \cite{zhu2021graph} instead of the pair-wise edge classifier. This removes the kernel selection process before training and significantly reduces training time. We use Eq. \ref{GBKProcess} to aggregate and update the embedding layers similarly. The formal definition of the gate generator is:
\begin{equation}
\small
    H(\mathbf{G})=\frac{\sum_{\left(v_{i}, v_{j}\right) \in \mathbf{E}} \mathbb{I}\left(y_{i}=y_{j}\right)}{|\mathbf{E}|}
\end{equation}
\begin{equation}
\small
\alpha_{ij}=
    \begin{cases}
    \max((1-\epsilon) \mathbb{I}(y_{i}=y_{j}),\epsilon) & \text{$y_i$, $y_j$ $\in$ $V_{train}$} \\
    H(\mathbf{G}) & \text{otherwise} 
    \end{cases}
\label{newGate} 
\end{equation}
\begin{equation}
\small
\mathcal{L}_{Fast\ Im-GBK}=\mathcal{L}_{im}
\label{LabelGCNloss}
\end{equation}
where the hyper-parameter $\epsilon$ will serve as the minimum similarity threshold, and $\mathcal{L}_{im}$ is aforementioned Class-Imbalance Handling loss.

\section{Experiments}\label{secexp}

We address four questions to enhance our understanding of the model: 
\textbf{RQ1:} How do Im-GBK and fast-Im-GBK models perform in comparison to baselines in node classification for an imbalanced scenario? \textbf{RQ2:} How is the performance efficiency of fast-Im-GBK compared to other baseline models? \textbf{RQ3:} What is the role of each component in our Im-GBK model, and do they positively impact the classification performance?

\subsection{Experiment Settings}
\label{expdata}
\textbf{\textit{Datasets}}. We conduct experiments on four datasets from PyTorch Geometric \cite{fey2019fast}, which are commonly used in graph neural network literature, to evaluate the efficiency of our proposed models for the node classification task when classes are imbalanced. Table \ref{Statistics} presents a summary of the datasets. In addition, using CiteSeer and Cora datasets, we generate two extreme instances in which each minority class with random selection has only five training examples.
\begin{table*}
    \caption{Statistics of the node classification datasets.}
    \begin{center}
		\begin{small}
			\begin{sc}
			    \resizebox{\textwidth}{!}{
                \begin{tabular}{l|c|c|c|c|c|c}
                \hline
                {} &  \bfseries Hom.Ratio &      \bfseries Imbalance Ratio &  \bfseries Nodes &   \bfseries Edges &  \bfseries Features &  \bfseries Classes \\
                \hline
                  CiteSeer \cite{sen2008collective} &      0.736 &   2.655 &   3327 &    9104 &      3703 &        6 \\
                  Cora \cite{sen2008collective}     &      0.810 &   4.544 &   2708 &   10556 &      1433 &        7 \\
                  Wiki \cite{sen2008collective}     &      0.712 &  45.111 &   2405 &   17981 &      4973 &       17 \\
                Coauthor CS \cite{shchur2018pitfalls}  &      0.808 &  35.051 &  18333 &  163788 &      6805 &       15 \\
                \hline
                \end{tabular}
                }
			\end{sc}
		\end{small}
	\end{center}
\label{Statistics} 
\end{table*}

\textbf{\textit{Experiment environment}}. For each dataset, we randomly select 60\% of total samples for training, 20\% for validation, and the remaining 20\% for testing. We use Adam optimizer and set learning rate $lr, weight\ decay$ as 0.001 and $5e^{-4}$, respectively. All baselines follow the same setting except for the layer number of 128. We run each experiment on the same random seeds to ensure reproducibility. Model training is done on NVIDIA GeForce RTX 3090 (24GB) GPU with 90GB memory. The code depends on PyTorch 1.7.0 and PyG 2.0.4.

\textbf{\textit{Evaluation metric}}. The quality of classification is assessed by average accuracy, AUC-ROC, and F1 scores. Each experiment is repeated five times to avoid randomness and compute the final value.

\textbf{\textit{Baselines}}. We evaluate our approach with representative and state-of-the-art approaches, including three classic GNN models (GCN \cite{kipf2017semisupervised}, GAT \cite{veličković2018graph}, GraphSage \cite{hamilton2018inductive}) using three traditional imbalance techniques (Over-sampling, Re-weight, SMOTE \cite{chawla2002smote}), original GBK-GNN \cite{2021GBK}, GraphSMOTE \cite{zhao2021graphsmote}, and TAM \cite{song2022tam} (we choose the combination of GCN+TAM+Balanced Softmax and GCN+TAM+ENS, referred to as GCN-TAM-BS and GCN-TAM-ENS, respectively).

\subsection{Comparisons with baselines (RQ1)}\label{rq1}

\begin{figure*}[t]
  {\includegraphics[width=0.95\linewidth]{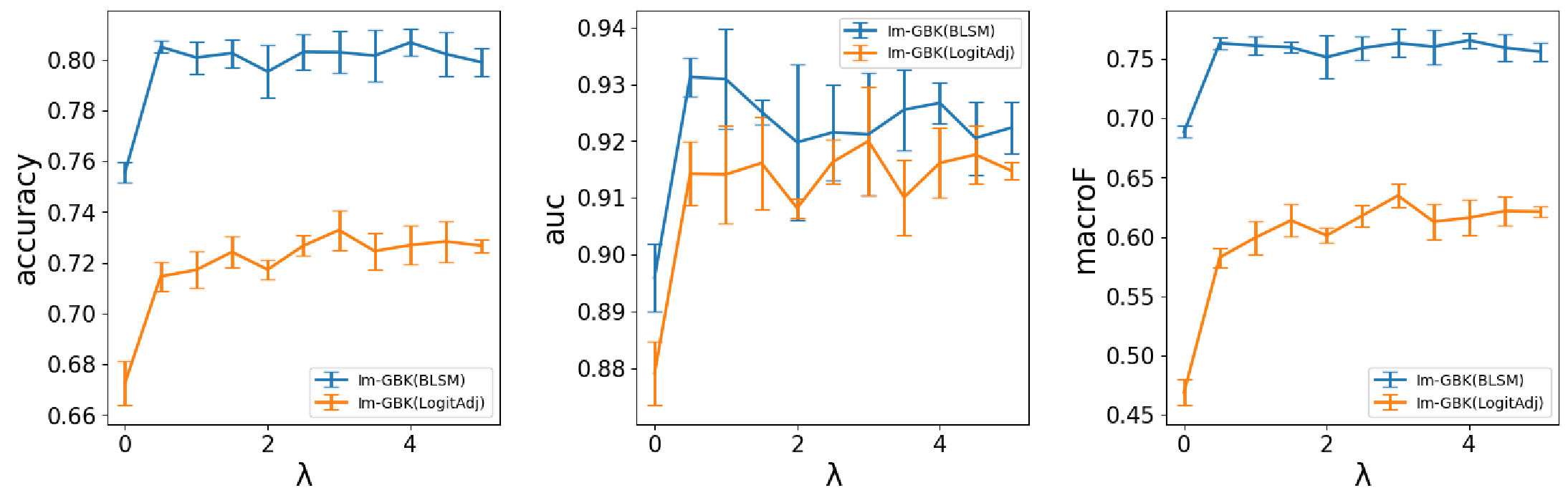}}
  \caption{Experimental results on Cora (extreme)}
  \label{fig:param}
\end{figure*}

In this section, we analyze the performance of classic graph neural networks (GNNs) and traditional imbalance learning approaches. As shown in the learning objective (Section \ref{losssubs}), $\lambda$ plays an important role in the tradeoff between classification error and consistency. Thus, we first explore the impact of hyperparameter $\lambda$ on the performance. We set $\lambda$ between 0 and 5 with an interval of 0.5 and the results are shown in Fig. \ref{fig:param}. According to our analysis, the impact of $\lambda$ on the results is insignificant if it is not 0. Therefore, in the following experiments, we always set $\lambda = 1$, considering the overall trends for both methods on all datasets. We examine the performance of different models. The results are reported in Table \ref{exp1} and Table \ref{exp2} on original datasets and extreme datasets, respectively. From the results, it can be observed that:
\begin{table*}[t]
    \caption{Comparison of different methods on original datasets.}
    \begin{center}
		\begin{small}
			\begin{sc}
			    \resizebox{\textwidth}{!}{
    \begin{tabular}{l|ccc|ccc|ccc|ccc}
    \hline
                             & \multicolumn{3}{c|}{Cora}                                                                              & \multicolumn{3}{c|}{CiteSeer}                                                                          & \multicolumn{3}{c|}{Wiki}                                                                              & \multicolumn{3}{c}{Coauthor-CS}                                                            \\ \hline
                             & \multicolumn{1}{c|}{ACC}            & \multicolumn{1}{c|}{AUC}            & F-1                        & \multicolumn{1}{c|}{ACC}            & \multicolumn{1}{c|}{AUC}            & F-1                        & \multicolumn{1}{c|}{ACC}            & \multicolumn{1}{c|}{AUC}            & F-1                        & \multicolumn{1}{c|}{ACC}            & \multicolumn{1}{c|}{AUC}            & F-1            \\ \hline
    GCN                      & \multicolumn{1}{c|}{0.853}          & \multicolumn{1}{c|}{0.981}          & 0.847                      & \multicolumn{1}{c|}{0.719}          & \multicolumn{1}{c|}{0.905}          & 0.687                      & \multicolumn{1}{c|}{0.664}          & \multicolumn{1}{c|}{0.875}          & 0.592                      & \multicolumn{1}{c|}{0.935}          & \multicolumn{1}{c|}{0.995}          & 0.914          \\ 
    GCN+SMOTE                & \multicolumn{1}{c|}{0.855}          & \multicolumn{1}{c|}{0.980}          & 0.848                      & \multicolumn{1}{c|}{0.709}          & \multicolumn{1}{c|}{0.904}          & 0.673                      & \multicolumn{1}{c|}{0.649}          & \multicolumn{1}{c|}{0.865}          & 0.605                      & \multicolumn{1}{c|}{0.939}          & \multicolumn{1}{c|}{0.996}          & 0.921          \\ 
    GCN+Re-weight            & \multicolumn{1}{c|}{0.848}          & \multicolumn{1}{c|}{0.981}          & 0.840                      & \multicolumn{1}{c|}{0.712}          & \multicolumn{1}{c|}{0.905}          & 0.683                      & \multicolumn{1}{c|}{0.672}          & \multicolumn{1}{c|}{0.873}          & 0.631                      & \multicolumn{1}{c|}{0.935}          & \multicolumn{1}{c|}{0.996}          & 0.915          \\ 
    GAT                      & \multicolumn{1}{c|}{0.853}          & \multicolumn{1}{c|}{0.969}          & 0.846                      & \multicolumn{1}{c|}{0.730}          & \multicolumn{1}{c|}{0.896}          & 0.702                      & \multicolumn{1}{c|}{0.243}          & \multicolumn{1}{c|}{0.643}          & 0.191                      & \multicolumn{1}{c|}{0.889}          & \multicolumn{1}{c|}{0.975}          & 0.822          \\ 
    GAT+SMOTE                & \multicolumn{1}{c|}{0.839}          & \multicolumn{1}{c|}{0.970}          & 0.822                      & \multicolumn{1}{c|}{0.716}          & \multicolumn{1}{c|}{0.876}          & 0.687                      & \multicolumn{1}{c|}{0.281}          & \multicolumn{1}{c|}{0.691}          & 0.250                      & \multicolumn{1}{c|}{0.370}          & \multicolumn{1}{c|}{0.721}          & 0.185          \\ 
    GAT+Re-weight            & \multicolumn{1}{c|}{0.827}          & \multicolumn{1}{c|}{0.965}          & 0.821                      & \multicolumn{1}{c|}{0.703}          & \multicolumn{1}{c|}{0.888}          & 0.677                      & \multicolumn{1}{c|}{0.125}          & \multicolumn{1}{c|}{0.580}          & 0.111                      & \multicolumn{1}{c|}{0.913}          & \multicolumn{1}{c|}{0.987}          & 0.891          \\ 
    GraphSAGE                & \multicolumn{1}{c|}{0.805}          & \multicolumn{1}{c|}{0.968}          & 0.780                      & \multicolumn{1}{c|}{0.697}          & \multicolumn{1}{c|}{0.895}          & 0.676                      & \multicolumn{1}{c|}{0.691}          & \multicolumn{1}{c|}{0.877}          & 0.578                      & \multicolumn{1}{c|}{0.905}          & \multicolumn{1}{c|}{0.988}          & 0.829          \\ 
    GraphSAGE+SMOTE          & \multicolumn{1}{c|}{0.798}          & \multicolumn{1}{c|}{0.971}          & 0.776                      & \multicolumn{1}{c|}{0.724}          & \multicolumn{1}{c|}{0.902}          & 0.702                      & \multicolumn{1}{c|}{0.693}          & \multicolumn{1}{c|}{0.884}          & 0.688                      & \multicolumn{1}{c|}{0.633}          & \multicolumn{1}{c|}{0.947}          & 0.397          \\ 
    GraphSAGE+Re-weight      & \multicolumn{1}{l|}{0.794}          & \multicolumn{1}{l|}{0.966}          & \multicolumn{1}{l|}{0.769} & \multicolumn{1}{l|}{0.703}          & \multicolumn{1}{l|}{0.888}          & \multicolumn{1}{l|}{0.677} & \multicolumn{1}{l|}{0.668}          & \multicolumn{1}{l|}{0.884}          & 0.563                      & \multicolumn{1}{c|}{0.898}          & \multicolumn{1}{c|}{0.986}          & 0.816          \\ \hline
    GraphSMOTE               & \multicolumn{1}{c|}{0.872}          & \multicolumn{1}{c|}{0.984}          & 0.864                      & \multicolumn{1}{c|}{\textbf{0.769}} & \multicolumn{1}{c|}{\textbf{0.929}} & \textbf{0.741}             & \multicolumn{1}{c|}{0.569}          & \multicolumn{1}{c|}{0.859}          & 0.440                      & \multicolumn{1}{c|}{0.940}           & \multicolumn{1}{c|}{0.996}          & 0.926          \\ 
    GCN-TAM                  & \multicolumn{1}{c|}{\textbf{0.878}} & \multicolumn{1}{c|}{0.931}          & \textbf{0.868}             & \multicolumn{1}{c|}{0.752}          & \multicolumn{1}{c|}{0.840}          & 0.727                      & \multicolumn{1}{c|}{0.651}          & \multicolumn{1}{c|}{0.823}          & 0.563                      & \multicolumn{1}{c|}{0.929}          & \multicolumn{1}{c|}{0.956}          & 0.914          \\                                             
    GBK-GNN-BS               & \multicolumn{1}{c|}{0.876}          & \multicolumn{1}{c|}{0.974}          & 0.866                      & \multicolumn{1}{c|}{0.730}          & \multicolumn{1}{c|}{0.915}          & 0.707                      & \multicolumn{1}{c|}{0.681}          & \multicolumn{1}{c|}{0.881}          & 0.611                      & \multicolumn{1}{c|}{0.936}          & \multicolumn{1}{c|}{\textbf{0.997}}          & 0.918          \\                                             
    Im-GBK (LogitAdj)        & \multicolumn{1}{c|}{0.866}          & \multicolumn{1}{c|}{0.979}          & 0.853                      & \multicolumn{1}{c|}{0.728}          & \multicolumn{1}{c|}{0.912}          & 0.699                      & \multicolumn{1}{c|}{0.674}          & \multicolumn{1}{c|}{0.887}          & 0.622                      & \multicolumn{1}{c|}{0.932}          & \multicolumn{1}{c|}{0.996}          & 0.912          \\ 
    Im-GBK (BLSM)           & \multicolumn{1}{c|}{0.861}           & \multicolumn{1}{c|}{0.979}          & 0.846                      & \multicolumn{1}{c|}{0.721}          & \multicolumn{1}{c|}{0.909}          & 0.700                      & \multicolumn{1}{c|}{0.677}          & \multicolumn{1}{c|}{0.888}          & 0.615                      & \multicolumn{1}{c|}{0.933}          & \multicolumn{1}{c|}{0.996}          & 0.914          \\ \hline
    Fast Im-GBK             & \multicolumn{1}{c|}{0.876}   & \multicolumn{1}{c|}{\textbf{0.988}} & 0.863                      & \multicolumn{1}{c|}{0.766}          & \multicolumn{1}{c|}{0.926}          & 0.738                      & \multicolumn{1}{c|}{\textbf{0.723}} & \multicolumn{1}{c|}{\textbf{0.911}} & \textbf{0.655}             & \multicolumn{1}{c|}{\textbf{0.951}} & \multicolumn{1}{c|}{\textbf{0.997}} & \textbf{0.939} \\ \hline
     \end{tabular}}
			\end{sc}
		\end{small}
	\end{center}
\label{exp1} 
\end{table*}

\begin{table*}[]
    \centering
    \small
    \caption{Comparison of different methods on extreme datasets.}\label{exp2}
    \noindent\makebox[\textwidth]{
    \begin{tabular}{l|ccc|ccc}
    \hline
                             & \multicolumn{3}{c|}{Cora Extreme}                                                                      & \multicolumn{3}{c}{CiteSeer Extreme}                                                                   \\ \hline
                             & \multicolumn{1}{c|}{ACC}            & \multicolumn{1}{c|}{AUC}            & F-1                        & \multicolumn{1}{c|}{ACC}            & \multicolumn{1}{c|}{AUC}            & F-1                        \\ \hline
    GCN                      & \multicolumn{1}{c|}{0.746}          & \multicolumn{1}{c|}{0.929}          & 0.647                      & \multicolumn{1}{c|}{0.697}          & \multicolumn{1}{c|}{0.877}          & 0.607                      \\ 
    GCN+SMOTE                & \multicolumn{1}{c|}{0.745}          & \multicolumn{1}{c|}{0.930}          & 0.648                      & \multicolumn{1}{c|}{0.699}          & \multicolumn{1}{c|}{0.877}          & 0.610                      \\ 
    GCN+Re-weight            & \multicolumn{1}{c|}{0.756}          & \multicolumn{1}{c|}{0.934}          & 0.667                      & \multicolumn{1}{c|}{0.700}          & \multicolumn{1}{c|}{0.878}          & 0.612                      \\ 
    GAT                      & \multicolumn{1}{c|}{0.679}          & \multicolumn{1}{c|}{0.878}          & 0.532                      & \multicolumn{1}{c|}{0.694}          & \multicolumn{1}{c|}{0.880}          & 0.605                      \\ 
    GAT+SMOTE                & \multicolumn{1}{c|}{0.653}          & \multicolumn{1}{c|}{0.861}          & 0.427                      & \multicolumn{1}{c|}{0.677}          & \multicolumn{1}{c|}{0.847}          & 0.593                      \\ 
    GAT+Re-weight            & \multicolumn{1}{c|}{0.689}          & \multicolumn{1}{c|}{0.869}          & 0.516                      & \multicolumn{1}{c|}{0.681}          & \multicolumn{1}{c|}{0.873}          & 0.596                      \\ 
    GraphSAGE                & \multicolumn{1}{c|}{0.657}          & \multicolumn{1}{c|}{0.873}          & 0.498                      & \multicolumn{1}{c|}{0.682}          & \multicolumn{1}{c|}{0.866}          & 0.596                      \\ 
    GraphSAGE+SMOTE          & \multicolumn{1}{c|}{0.671}          & \multicolumn{1}{c|}{0.884}          & 0.494                      & \multicolumn{1}{c|}{0.680}          & \multicolumn{1}{c|}{0.871}          & 0.593                      \\ 
    GraphSAGE+Re-weight      & \multicolumn{1}{c|}{0.674}          & \multicolumn{1}{l|}{0.878}          & \multicolumn{1}{l|}{0.509} & \multicolumn{1}{c|}{0.687}          & \multicolumn{1}{l|}{0.873}          & \multicolumn{1}{l}{0.598}  \\ 
    GraphSMOTE               & \multicolumn{1}{c|}{0.770}          & \multicolumn{1}{c|}{0.928}          & 0.674                      & \multicolumn{1}{c|}{0.703}          & \multicolumn{1}{c|}{0.893}          & 0.612                      \\ 
    GCN-TAM-BS               & \multicolumn{1}{c|}{\textbf{0.829}} & \multicolumn{1}{c|}{0.888}          & \textbf{0.797}             & \multicolumn{1}{c|}{0.718}          & \multicolumn{1}{c|}{0.801}           & 0.649                      \\ 
    GCN-TAM-ENS              & \multicolumn{1}{c|}{0.790}  & \multicolumn{1}{c|}{0.884}          & {0.769}             & \multicolumn{1}{c|}{0.680}          & \multicolumn{1}{c|}{0.797}           & \textbf{0.659}                      \\ 
    GBK-GNN                  & \multicolumn{1}{c|}{0.692}          & \multicolumn{1}{c|}{0.905}          & 0.521                      & \multicolumn{1}{c|}{0.696}          & \multicolumn{1}{c|}{0.892}           & 0.607                      \\ \hline
    Im-GBK (LogitAdj)        & \multicolumn{1}{c|}{0.717}          & \multicolumn{1}{c|}{0.914}          & 0.599                      & \multicolumn{1}{c|}{0.703}          & \multicolumn{1}{c|}{0.896}          & 0.615                      \\ 
    Im-GBK (BLSM)            & \multicolumn{1}{c|}{0.800}          & \multicolumn{1}{c|}{0.931}          & 0.761                      & \multicolumn{1}{c|}{0.705}          & \multicolumn{1}{c|}{0.897}          & {0.655}             \\ \hline
    Fast Im-GBK              & \multicolumn{1}{c|}{0.727}          & \multicolumn{1}{c|}{\textbf{0.941}} & 0.570                      & \multicolumn{1}{c|}{\textbf{0.737}} & \multicolumn{1}{c|}{\textbf{0.899}} & 0.641                      \\ \hline
    \end{tabular}}
     
    \end{table*}

\begin{compactitem}
    \item The experiment demonstrates that Im-GBK models, which use logit adjust loss and balanced softmax (denoted as Im-GBK (LogitAdj) and Im-GBK (BLSM), respectively), achieve comparable or better results to state-of-the-art methods in most original datasets. Specifically, in Cora and CiteSeer, Im-GBK (LogitAdj) and Im-GBK (BLSM) are comparable to GraphSMOTE and GCN-TAM, while they outperform it on Wiki. Furthermore, Fast Im-GBK demonstrates superiority on Wiki and Coauthor-CS, and all proposed methods performed better than the baselines in extremely imbalanced cases.
    \item The models designed specifically for imbalance problems perform better in the extreme cases (Cora Extreme and CiteSeer Extreme); refer to Table \ref{exp2}. Our models show superior performance in extreme circumstances, as indicated by achieving the best performance consistently w.r.t the most of metrics, whereas GCN-TAM and GraphSMOTE, as specifically designed models for the imbalanced classification task, also exhibit their capabilities in differentiating minority classes.
    \item Models designed to account for graph heterophily generally outperform classic GNNs and modified classic GNNs. This also empirically validates the rationality of using heterophilic neighborhoods for imbalanced classification to some extent.
\end{compactitem}

Overall, the experiment demonstrates that the proposed methods are effective in imbalanced node classification and offer better or comparable performance to state-of-the-art. In extreme cases, our approaches outperform and show a better capability in differentiating minority classes.

\subsection{Comparison in Efficiency (RQ2)}
\label{rq2}

Section \ref{mt} showed that the proposed Im-GBK model could be time-consuming due to its second loss function component. To address this, we replace the kernel selection process by using a graph-level homophily ratio. Table \ref{processingTime} presents training time comparisons, revealing that models using gate-selection mechanisms, like GBK-GNN and Im-GBK, require significantly more training time. 
For example, while most models could complete one epoch within 1 second, GBK-GNN and Im-GBK took over 10 seconds and around 160 seconds to train one epoch on the CS dataset. 
GraphSMOTE also requires more time than Fast Im-GBK because it has to generate several synthetic nodes for each minority class. However, the results demonstrate that our proposed Fast Im-GBK model shows a significant reduction in training time compared to GBK-GNN and GraphSMOTE. 

\subsection{Ablation Analysis (RQ3)}

\begin{wraptable}{R}{0.34\textwidth}
\caption{Average Execution Time (s) per epoch on CS.}
\small
\begin{tabular}{l|c}
\hline
{} &       Time \\
\hline
 GCN                 &       0.0166 \\
 GAT                 &       0.1419 \\
 GraphSage           &       0.0135 \\
 GraphSMOTE          &       5.309 \\
 Fast Im-GBK         &       0.5897 \\
 GBK-GNN             &       12.320\\
 Im-GBK(LogitAdj)    &       11.594\\
 Im-GBK(BLSM)        &       11.271\\
 \hline
 \end{tabular}
\label{processingTime}
\end{wraptable}
Subsection \ref{rq2} showed that the proposed method exhibits clear advantages in performance compared to other baselines in differentiating minority classes for extremely imbalanced graphs. To further investigate the fundamental factors underlying the performance improvements of our proposed method in Im-GBK, we conduct ablation analyses using one of the extreme cases, CiteSeer Extreme. We show the effectiveness of the model handling class imbalance classification by ablating the model Class-Imbalance Handler and Heterophily Handler, respectively. 
In Table \ref{abtb}, `Class-Imbalance Handling Loss' represents two Class-Imbalance Handling losses introduced in Section \ref{m:logit}. `Heterophily handling' refers to the method introduced in Section \ref{m:HH} to capture graph heterophily, and `$\times$' means this part is ablated. Considering all strategies, it can be observed from Table \ref{abtb} that either dropping `Class-Imbalance Handling Loss' or `Heterophily handling' components will result in a decrease in performance.

\begin{table}[t]
    \centering
    \small
    \caption{Ablation Experiment Results}
    \begin{tabular}{cc|c|c|c|c}
        \hline
        \multicolumn{2}{c|}{Class-Imbalance Handling Loss}                 & \multirow{2}{*}{\begin{tabular}[c]{@{}c@{}}Heterophily\\ Handling\end{tabular}} & \multirow{2}{*}{ACC} & \multirow{2}{*}{AUC} & \multirow{2}{*}{F1} \\ \cline{1-2}
        \multicolumn{1}{c|}{Logit adjusted loss}   & Balanced Softmax      &                                                                                 &                      &                      &                     \\ \hline
        \multicolumn{1}{c|}{$\times$} & $\times$ & $\times$        & 0.697                    & 0.877                    & 0.607                   \\ 
        \multicolumn{1}{c|}{$\times$} & $\times$ & $\surd$        & 0.696                    & 0.892                    & 0.607                   \\ 
        \multicolumn{1}{c|}{$\times$} & $\surd$ & $\times$        & \textbf{0.710}           & 0.871                    & 0.618                   \\ 
        \multicolumn{1}{c|}{$\times$} & $\surd$ & $\surd$        & 0.705                    & \textbf{0.897}           & \textbf{0.655}          \\ 
        \multicolumn{1}{c|}{$\surd$} & $\times$ & $\times$        & 0.667                    & 0.872                    & 0.566                   \\ 
        \multicolumn{1}{c|}{$\surd$} & $\times$ & $\surd$        & 0.703                    & 0.896                    & 0.615                   \\  \hline
        \end{tabular}
        \label{abtb}
    \end{table}

\section{Conclusion}
In this paper, we studied the problem of imbalanced classification on graphs from the perspective of graph heterophily. We observed that if a model cannot handle heterophilic neighborhoods in graphs, its ability to address imbalanced classification will be impaired. To address the graph imbalance problem effectively, we proposed a novel framework, Im-GBK, and its faster version, Im-GBK, that simultaneously tackles heterophily and class imbalance. Our framework overcomes the limitations of previous techniques by achieving higher efficiency while maintaining comparable performance. Extensive experiments are conducted on various real-world datasets, demonstrating that our model outperforms most baselines. Furthermore, a comprehensive parameter analysis is performed to validate the efficacy of our approach. In future research, we aim to explore alternative methods for modeling graph heterophily and extend our approach to real-world applications, such as fraud and spammer detection.









\bibliographystyle{spmpsci}
\bibliography{refs} 

\begin{thebibliography}{10}
\providecommand{\url}[1]{{#1}}
\providecommand{\urlprefix}{URL }
\expandafter\ifx\csname urlstyle\endcsname\relax
  \providecommand{\doi}[1]{DOI~\discretionary{}{}{}#1}\else
  \providecommand{\doi}{DOI~\discretionary{}{}{}\begingroup \urlstyle{rm}\Url}\fi

\bibitem{chawla2002smote}
Chawla, N.V., Bowyer, K.W., Hall, L.O., Kegelmeyer, W.P.: Smote: synthetic minority over-sampling technique.
\newblock Journal of artificial intelligence research \textbf{16}, 321--357 (2002)

\bibitem{2021GBK}
Du, L., Shi, X., Fu, Q., Ma, X., Liu, H., Han, S., Zhang, D.: Gbk-gnn: Gated bi-kernel graph neural networks for modeling both homophily and heterophily.
\newblock In: Proceedings of the ACM Web Conference 2022, pp. 1550--1558 (2022)

\bibitem{fey2019fast}
Fey, M., Lenssen, J.E.: Fast graph representation learning with pytorch geometric.
\newblock arXiv preprint arXiv:1903.02428  (2019)

\bibitem{hamilton2018inductive}
Hamilton, W.L., Ying, R., Leskovec, J.: Inductive representation learning on large graphs (2018)

\bibitem{japkowicz2002class}
Japkowicz, N., Stephen, S.: The class imbalance problem: A systematic study.
\newblock Intelligent data analysis \textbf{6}(5), 429--449 (2002)

\bibitem{johnson2019survey}
Johnson, J.M., Khoshgoftaar, T.M.: Survey on deep learning with class imbalance.
\newblock Journal of Big Data \textbf{6}(1), 1--54 (2019)

\bibitem{kipf2017semisupervised}
Kipf, T.N., Welling, M.: Semi-supervised classification with graph convolutional networks (2017)

\bibitem{liu2021pick}
Liu, Y., Ao, X., Qin, Z., Chi, J., Feng, J., Yang, H., He, Q.: Pick and choose: a gnn-based imbalanced learning approach for fraud detection.
\newblock In: Proceedings of the Web Conference 2021, pp. 3168--3177 (2021)

\bibitem{liu2022towards}
Liu, Y., Zheng, Y., Zhang, D., Chen, H., Peng, H., Pan, S.: Towards unsupervised deep graph structure learning.
\newblock In: Proceedings of the ACM Web Conference 2022, pp. 1392--1403 (2022)

\bibitem{mcpherson2001birds}
McPherson, M., Smith-Lovin, L., Cook, J.M.: Birds of a feather: Homophily in social networks.
\newblock Annual review of sociology \textbf{27}(1), 415--444 (2001)

\bibitem{longtail}
Menon, A.K., Jayasumana, S., Rawat, A.S., Jain, H., Veit, A., Kumar, S.: Long-tail learning via logit adjustment (2021)

\bibitem{park2021graphens}
Park, J., Song, J., Yang, E.: Graphens: Neighbor-aware ego network synthesis for class-imbalanced node classification.
\newblock In: International Conference on Learning Representations (2021)

\bibitem{ren2020balanced}
Ren, J., Yu, C., Sheng, S., Ma, X., Zhao, H., Yi, S., Li, H.: Balanced meta-softmax for long-tailed visual recognition (2020)

\bibitem{sen2008collective}
Sen, P., Namata, G., Bilgic, M., Getoor, L., Galligher, B., Eliassi-Rad, T.: Collective classification in network data.
\newblock AI magazine \textbf{29}(3), 93--93 (2008)

\bibitem{shchur2018pitfalls}
Shchur, O., Mumme, M., Bojchevski, A., G{\"u}nnemann, S.: Pitfalls of graph neural network evaluation.
\newblock arXiv preprint arXiv:1811.05868  (2018)

\bibitem{song2022tam}
Song, J., Park, J., Yang, E.: Tam: Topology-aware margin loss for class-imbalanced node classification.
\newblock In: International Conference on Machine Learning, pp. 20,369--20,383. PMLR (2022)

\bibitem{veličković2018graph}
Veličković, P., Cucurull, G., Casanova, A., Romero, A., Liò, P., Bengio, Y.: Graph attention networks (2018)

\bibitem{wu2020graph}
Wu, Y., Lian, D., Xu, Y., Wu, L., Chen, E.: Graph convolutional networks with markov random field reasoning for social spammer detection.
\newblock In: Proceedings of the AAAI conference on artificial intelligence, vol.~34, pp. 1054--1061 (2020)

\bibitem{zhao2021graphsmote}
Zhao, T., Zhang, X., Wang, S.: Graphsmote: Imbalanced node classification on graphs with graph neural networks.
\newblock In: Proceedings of the 14th ACM international conference on web search and data mining, pp. 833--841 (2021)

\bibitem{zheng2022graph}
Zheng, X., Liu, Y., Pan, S., Zhang, M., Jin, D., Yu, P.S.: Graph neural networks for graphs with heterophily: A survey.
\newblock arXiv preprint arXiv:2202.07082  (2022)

\bibitem{zhu2021graph}
Zhu, J., Rossi, R.A., Rao, A., Mai, T., Lipka, N., Ahmed, N.K., Koutra, D.: Graph neural networks with heterophily.
\newblock In: Proceedings of the AAAI Conference on Artificial Intelligence, vol.~35, pp. 11,168--11,176 (2021)

\bibitem{zhu2020beyond}
Zhu, J., Yan, Y., Zhao, L., Heimann, M., Akoglu, L., Koutra, D.: Beyond homophily in graph neural networks: Current limitations and effective designs.
\newblock Advances in Neural Information Processing Systems \textbf{33}, 7793--7804 (2020)

\end{thebibliography}
\end{document}